%% file: main.tex
\pdfoutput=1

\documentclass[11pt]{article}

\usepackage{acl}

\usepackage{times}
\usepackage{latexsym}
\usepackage{microtype}
\usepackage{graphicx}
\usepackage{adjustbox}
\usepackage{multirow}
\usepackage{booktabs}
\usepackage{tabularx}
\usepackage[inline]{enumitem}
\usepackage{amsmath}
\usepackage{alphabeta}
\usepackage{flushend}
\usepackage{balance}
\usepackage[T1]{fontenc}

\usepackage[utf8]{inputenc}

\usepackage[bottom]{footmisc}
\usepackage[section]{placeins}

\title{Discovering Fine-Grained Semantics in Knowledge Graph Relations} 

\author{Nitisha Jain \\
  Hasso Plattner Institute\\
  University of Potsdam \\
 Potsdam, Germany \\
  \texttt{nitisha.jain@hpi.de} \\
  \And
  Ralf Krestel\\
  ZBW - Leibniz Centre for Economics \& \\
  Kiel University\\
  Kiel, Germany \\
  \texttt{r.krestel@zbw.eu}\\}

\begin{document}
\maketitle

\begin{abstract}
When it comes to comprehending and analyzing multi-relational data, the semantics of relations are crucial. Polysemous relations between different types of entities, that represent multiple semantics, are common in real-world relational datasets represented by knowledge graphs. For numerous use cases, such as entity type classification, question answering and knowledge graph completion, the correct semantic interpretation of these relations is necessary. In this work, we provide a strategy for discovering the different semantics associated with abstract relations and deriving many sub-relations with fine-grained meaning. To do this, we leverage the types of the entities associated with the relations and cluster the vector representations of entities and relations. The suggested method is able to automatically discover the best number of sub-relations for a polysemous relation and determine their semantic interpretation, according to our empirical evaluation.
\end{abstract}

\section{Introduction}
\label{sec:introduction}

Relations between different words or phrases are important for the semantic understanding of text. Popular knowledge graphs (KGs) such as Yago~\cite{mahdisoltani2014yago3}, NELL~\cite{mitchell2018never} and DBpedia~\cite{dbpedia} are formulated in terms of relational databases where the entities are linked to each other with different relations or predicates. In real-world textual data, the relations are often polysemous by nature, i.e., they exhibit distinct meanings in different contexts. Similar to the task of word sense disambiguation, that is required to understand different contextual meanings of words, relation disambiguation is needed to interpret the specific, contextual semantics of relations. Relation polysemy occurs frequently in open texts and has been studied and discussed by previous works on automatic relation extraction from texts~\cite{min2012ensemble,galarraga2014canonicalizing,han2016global}.
Relation semantics are particularly important in the context of knowledge graphs which are widely used for systematic representation of data in the form of $\langle subject, predicate, object \rangle$ triples. Here, \textit{subject} and \textit{object} are chosen from a set of entities, while the \textit{predicate} that links the entities to each other belongs to a set of relations. As the triples in KGs are derived from real-world facts, ambiguity from texts often makes it way into the KG triples as well. 
Specifically, the relations may represent multiple meanings depending on the context, which in the case of KG triples, is defined by the types of the entities being connected by the relations.
The underlying idea of defining the role of words by their context is quite old in Linguistics, advocated by Firth in 1957 as `a word is characterized by the company it keeps'~\cite{firth1957synopsis}. In the context of KGs, one could say \textit{`a relation is characterized by the entity types it connects'}.

In order to gauge the issue of multiple relation semantics in popular KGs, we analysed the relations in the Yago3 dataset in terms of the number of unique entity types pairs that were found in the associated triples. The results are plotted in Figure~\ref{fig:YagoRelationClassTypes}. 
It can be seen that the majority of the relations have multiple entity types associated with them, with generic relations such as $owns$ and $created$ exhibiting very high plurality. Similar insights were also derived from the NELL knowledge graph. Some examples of the actual entity types associated with polysemous relations from these KGs 
are shown in Table~\ref{tbl:sampletypes}.

\input{figures/relations_plots}

In this work we advocate that for such relations that are associated with a number of different entity type pairs, 
it would be prudent to instead split them and create new sub-relations that have a more concise meaning according to the context. The meanings of the sub-relations could be identified based on the distinct types of the associated entities and therefore, be clearly defined.
However, this approach can be tricky since entity types can vary widely. 
While some types such as \textit{television} and \textit{movie} for the \textit{created} relation in Yago are semantically similar to one another, other types are quite different, for instance \textit{company} and \textit{writer}, or \textit{airport} and \textit{club} in the case of \textit{owns} relation. 
Due to the complex hierarchy of classes (entity types)\footnote{the terms \textit{class} and \textit{entity type} will be used interchangeably in the paper from this point on} in the underlying schema (or ontology), entity types belong to different granularity levels~\cite{jain2021embeddings} %
leading to a wide range of semantic similarity between them. 
It is, therefore, a non-trivial task to decide how a relation should be split into sub-relations based on the semantics of the entity types associated with it, both in terms of the number of sub-relations as well the subset of entity types that the sub-relations should encompass. %

In this work, we define the problem statement of \textit{fine-gained relation refinement} which concerns with the disambiguation of polysemous relations in relational datasets and 
propose a data-driven and scalable method named \textit{\textbf{FineGReS}} \textit{(\textbf{F}ine-\textbf{G}rained \textbf{R}elation \textbf{S}emantics)} as a first solution towards the same. 
The fine-grained semantics for the relations are derived by relying on the multiple semantics of the relations evident from the types of the associated entities.  
The proposed approach leverages knowledge graph embeddings, that provide representations of the KG entities and relations in a continuous vector space. We find optimal clusters in the latent space to identify the underlying semantic features such that polysemous relations can be represented in terms of multiple sub-relations with well-defined semantics. This approach is able to automatically determine not only the optimal number of sub-relations (corresponding to the number of clusters), but also the entity types that should be associated with each of them so as to have the best semantic representation.
Experimental evaluation performed on popular relational datasets reveals the benefits of defining fine-grained semantics and brings forth the efficacy of the proposed approach in the face of the challenging nature of this task.

\input{tables/examplesemantics}

\paragraph{Contributions.}
\begin{enumerate*}
\item We formally define the task of \textit{fine-grained relation refinement} in relational datasets and motivate its importance and benefits. 
\item We propose a data-driven and scalable method  \mbox{\textit{FineGReS}} to identify multiple sub-relations that capture the different semantics of the relations via clustering in the latent space and empirically evaluate their quality. 
\item Additionally, we illustrate the benefits achieved by substituting a polysemous relation with multiple fine-grained sub-relations as obtained from the proposed approach on downstream applications, such as entity classification.
\end{enumerate*}
The code and data will be made publicly available.

\section{Fine-Grained Relation Semantics}

Relation polysemy is quite common in knowledge graphs due to two primary reasons. 
Firstly, the schema for most large scale KGs that are in use today have been constructed through manual or semi-automated efforts, where the relations between the entities are curated from text.
Relations are often abstracted in such KGs for simplification and avoidance of redundancies. This may result in cases where a single relation serves as a general notion between various different types of KG entities and has more than one semantic meaning associated with it. %
In addition to this, the fact that these KGs represent real-world facts that are expressed in natural language having inherent ambiguities, contributes further to the relation polysemy in KGs. For instance, the relation phrase `\textit{part of}' represents varied semantics based on its context of biology (\textit{finger part of hand}), organizations (\textit{Google part of Alphabet}), geography (\textit{Amazon part of South America}) and many others. Even in KGs that have a large number of different relations can suffer from ambiguous relations, for instance DBpedia has around 300 relations that are relatively well-defined in terms of their entity types, however there exist relations such as \textit{award} and \textit{partOf} that still convey ambiguity.
The determination of fine-grained relation semantics in relational data is an important task which can bring substantial benefits to a wide range of NLP and semantic use cases as discussed further in this section.

The task of \textbf{relation extraction} is essential for information extraction from texts and it continues to be challenging due to the varied semantics of the evolving language. 
For identifying patterns and extracting relation mentions from text, unsupervised techniques typically rely on the predefined types of relation arguments~\cite{hasegawa2004discovering,shinyama2006preemptive,chen2005unsupervised}. 
Given an existing KG and schema, with the goal to extract facts for a particular relation from a new corpus of text, a distant supervision approach will leverage relation patterns based on the types of entities over the text. As an example, if the relation \textit{created} has been established between a \textit{painter} and \textit{artwork}, then the identification of this relation can be aided by specific patterns in text. However, if the relation \textit{created} is generically defined between any \textit{person} entity and any \textit{work} entity, then the resulting text patterns for this relation will be noisy and varied, therefore may fail to identify the correct fact triples from text. Identifying the different meanings of a relation in different contexts can help with defining concrete patterns for extraction of relation phrases.

This is also useful for identification and \textbf{classification of entities} by their types in a knowledge graph. E.g. the target entity of the relation \textit{directed} is likely to be of type \textit{movie} or \textit{play}. If the relations have a wider semantic range, the type of entities cannot be identified at a fine-grained level. For instance, it might be only possible to identify the entity type as \textit{work} and not specifically \textit{movie}, which could adversely affect the performance of further applications such as \textbf{entity linking} and \textbf{question answering}. Numerous question answering systems that use knowledge graphs as back-end data repositories (KBQA)~\cite{cui2019kbqa} rely on the type information of the entities to narrow down the search space for the correct answers. Thus, distinct relation semantics in terms of the types of connected entities are essential for supporting QA applications over KGs.

We would like to emphasize that task of defining fine-grained relation semantics is important in the context of \textbf{KG refinement}, not being merely limited to already existing datasets but in general.
KGs usually evolve over time and often in a fragmented fashion, where new facts might be added to a KG that do not strictly conform or can be correctly encapsulated by the existing ontology. Addition of such new facts might easily lead to noisy and abstracted semantics in previously well-defined KG relations.  
Relation disambiguation would therefore play a important role in identifying new fine-grained sub-relations with precise semantics.
The proposed \textit{FineGReS} method is generally applicable and could prove to be incredibly useful in all the above scenarios.  

From these examples,
it is therefore evident that fine-grained relation semantics can be beneficial to a wide variety of use cases and is indeed an important problem that has not been fully explored yet. In the next section, we provide the necessary background and present the problem statement.

\section{Preliminaries}
\label{sec:prelimimaries}

\paragraph{Knowledge Graph.}
For a knowledge graph $\mathcal{G}$, the set of unique relations %
is denoted as $\mathcal{R}$. A KG fact (or triple) $F$ = $\langle e_h, r, e_t \rangle$ consists of the head entity $e_h$, the tail entity $e_t$ and the relation $r$ that connects them, where $e_h$ and $e_t$ belong to the set of entities $\mathcal{E}$. Any given relation $r$ $\in$ $\mathcal{R}$ appears in several triples, forming a subset $\mathcal{G'}$ of $\mathcal{G}$. 

\paragraph{Entity Types.}
The semantic types or classes of the entities are defined in an ontology associated with a KG that defines its schema. The entities $e$ $\in$ $\mathcal{E}$ are connected with their types by ontological triples such as $\langle e, typeOf, t \rangle$, where $t \in \Tau$, the set of entity types in the ontology. 

We define a type pair as the tuple $\langle t_{h}, t_{t} \rangle$ where $\langle e_h, typeOf, t_{h}\rangle$ and $\langle e_t,typeOf, t_{t}\rangle$.
A set of unique type pairs (for a given relation $r$ and corresponding $\mathcal{G'}$ is denoted as $P_r$.
Thus we have,  
$P_r = \{\langle t_{h}, t_{t} \rangle | \langle e_h, typeOf, t_{h} \rangle, \langle e_t, typeOf, t_{t} \rangle, \langle e_h,r,e_t \rangle \in \mathcal{G'} \}$.  The total number of such unique type pairs for relation $r$, i.e. $|P_r|$ is denoted by $\mathcal{L}_{r}$.

\paragraph{KG Embeddings.}
\label{par:embeddings}

Knowledge graph embeddings have gained immense popularity and success for representation learning of relational data. They provide an efficient way to capture latent semantics of the entities and relations in KGs.  
The main advantage of these techniques is that they enable easy manipulation of KG components when represented as vectors in low dimensional space.  
For example in TransE~\cite{bordes2013translating}, for a triple $\langle h, r, t \rangle$ the vectors \textbf{h}, \textbf{r} and \textbf{t} satisfy the relation \textbf{h} + \textbf{r} = \textbf{t} or \textbf{r} = \textbf{t} - \textbf{h}. 
In this work, we leverage the representational abilities of the embeddings to obtain the semantic vectors for relations expressed in terms of the entities associated with them. For vectors \textbf{h}, \textbf{r} and \textbf{t} as obtained from an embedding corresponding to a KG triple $\langle e_h, r, e_t \rangle$, we define a vector \mbox{\boldmath$\Delta_{r_{i}}$} which is a function of \mbox{\boldmath$t_{i}$} and \mbox{\boldmath$h_{i}$}. 
Every \mbox{\boldmath$\Delta_{r_{i}}$} vector is mapped to a type pair $P_{r_{i}}$ as per the entities $e_h$, $e_t$ that they are both derived from.

\paragraph{Problem Definition.} 
Given a relation $r$ $\in$ $\mathcal{R}$ in $\mathcal{G}$, the set of \mbox{\boldmath$\Delta_{r}$} vectors and the corresponding set of type pairs $P_r$, the goal is to find for this relation an optimal configuration of clusters $\mathcal{C}_{opt}$ = $\{\mathcal{C}_1, \mathcal{C}_2 ... \mathcal{C}_N\}$, 
where the \mbox{\boldmath$\Delta_{r_{i}}$} vectors are uniquely distributed among the clusters i.e. each \mbox{\boldmath$\Delta_{r_i}$} $\in$ $\mathcal C_j$, $i = 1...|\mathcal{G'}|$, $j = 1...N$, s.t. an objective function $\mathcal{F}(\mathcal{C}_{opt})$ is maximized. 

Further, each cluster $\mathcal C_j$ represents the semantic union of the subset of type pairs $P'_r$ where \mbox{$\exists$ \boldmath{$\Delta_{r_{i}}$}} $\in$ $\mathcal C_j$ s.t. \mbox{\boldmath$\Delta_{r_{i}}$} is mapped to one of the type pairs in $P'_r$. 
Thus, the optimal configuration of clusters corresponds to the optimal number of sub-relations and their fine-grained semantics as defined by the type pairs that they represent. 
The proposed \mbox{\textit{FineGReS}} method can derive this optimal configuration for the relations of a KG.

\section{Method}
\label{sec:method}
In this section, we describe in detail the design and implementation details of the proposed \mbox{\textit{FineGReS}} method for a relation that can easily scaled to any number of relations in the dataset.

\subsection{Semantic Mapping for Facts}
\label{sec:datapoints}

For every unique relation $r$ in $\mathcal{G}$, we firstly find the subset of facts $\mathcal{G'}$ where $r$ appears. 
To understand the semantics of the entities associated with $r$,
the entities are mapped to their corresponding classes as defined in the underlying ontology.  
By doing so, we obtain a list of entity type pairs $\langle t_{h}, t_{t} \rangle$ for the relation.   
Note that several entities in $\mathcal{G'}$ might map to the same type and therefore, a single type pair tuple would be obtained several times.
Therefore in the next step, we identify the unique type pairs for a relation $r$ as the set $P$. At this stage, every fact in $\mathcal{G'}$ is associated with a type pair $\langle t_{h}, t_{t} \rangle \in P$ that represents the semantics of this fact. 
For example, for the $created$ relation, a triple $\langle DaVinci, created, Mona Lisa \rangle$ would be mapped to $\langle artist, painting\rangle$ as per the types of the head and tail entities.

\subsection{Vector Representations for Relations}
For representing the semantics of $r$ in terms of the associated entities, we leverage pre-trained KG embeddings. As proposed in previous work~\cite{jiang2020learning}, we obtain a representation for $r$ from \textbf{h} and \textbf{t} corresponding to every fact in $\mathcal{G'}$ and denote this vector as \mbox{\boldmath$\Delta_{r}$}.
In this way, for every relation $r$, a set of \mbox{\boldmath$\Delta_{r}$} vectors is obtained from the KG embeddings, in addition to the actual \textbf{r} vector that the embedding already provides. 
These \mbox{\boldmath$\Delta_{r}$} vectors are already mapped to the unique type pairs $P_i$
$\in$ $P$ for the relation $r$ (according to the fact triples they were calculated from)\footnote{we denote $P_r$ as $P$ when the relation $r$ is clear from the context}, such that each unique type pair is represented by a set of \mbox{\boldmath$\Delta_{r}$} vectors. 
These \mbox{\boldmath$\Delta_{r}$} vectors encode the information conveyed by both the head and tail entity types together and represent the relationship between the entities, they therefore represent the latent semantics of the relations in different facts. 
The \mbox{\boldmath$\Delta_{r}$} vectors become our data points (with the associated type pairs ${P_i}$ serving as their labels). 

\input{figures/vector_plots}

\paragraph{Relation Semantics.}
While it is believed that KG embeddings are able to capture relation similarity in the embedding space, i.e., relations having similar semantics occur close together in the vector space~\cite{transF2018knowledge,kalo2019knowledge}, %
we found that relations having multiple semantics 
(based on the context of their entities)
are, in fact, not represented well in the vector space.
In fact, for polysemous relations, the vectors obtained for a 
single relation (from the different facts that it appears in) form separate clusters in the vector space that do not overlap with the actual relation vector \textbf{r} obtained from the embeddings. 
This happens due to the fact that multiple entity pairs connected by the same relation are semantically different from one another.
Figures~\ref{fig:nell-vectors} illustrates an example from the NELL dataset where this behaviour of the embedding vectors for relations is clearly visible. 
We leverage this semantically-aware behaviour of the embedding vectors
to determine meaningful clusters of \mbox{\boldmath$\Delta_{r}$} vectors that represent the distinct latent semantics exhibited by different entity type pairs connected by the same relation, as described next.

\subsection{Clustering for Fine-grained Semantics}

For each relation $r$, the total number of unique type pairs $\mathcal{L}$ = $|P_r|$ is theoretically the maximum number of possible semantic sub-relations or clusters that could be obtained for $r$. This is the \textit{maximal} splitting that will assign a different sub-relation for every different type pair. 
However, in practice, it is rare that all the type pairs would have completely different semantics. 
For example, the $created$ relation in Yago has type pairs $\langle artist, painting\rangle$ and $\langle artist, music\rangle$ that have the same head entity type, while the type pair $\langle organization, software\rangle$ conveys quite a different meaning.
While a single relation is not sufficient to be representative of the semantics of all facts which is it appears in, at the same time, a naive \textit{maximal} splitting of the relation as per the unique type pairs would also be inefficient and lead to a large number of unnecessary sub-relations. 

The \textit{FineGReS} method aims to find an optimal number and composition of clusters $\mathcal{C}_{opt}$ for the type pairs that can convey distinct semantics of the relations based on the data, by combining similar type pairs while separating the dissimilar ones. Each of the clusters having one or more than one semantically similar type pairs represents a potential sub-relation. 
In order to obtain this configuration, various compositions of the clusters need to be analysed for optimality. Clustering is performed in an iterative manner with a predefined number of clusters and combinations of type pairs within each cluster. 
Since it is not feasible or practical to consider all possible clusters of the type pairs, \textit{FineGReS} leverages the semantic similarity of type pairs to narrow down the search space for obtaining the optimal clusters. For this, the similarity scores between all combinations of the unique type pairs $(t_{h_{i}}, t_{t_{i}})$, $(t_{h_{j}}, t_{t_{j}})$ are calculated. 
First, the vector representations for the types are obtained. Subsequently, the similarity scores are obtained by calculating the similarity scores between the vectors corresponding to $t_{h_{i}}$ and $t_{h_{j}}$ as well as $t_{t_{i}}$ and $t_{t_{j}}$ and then taking their mean value.

\paragraph{Iterative Clustering.}
The clustering begins with $\mathcal{L}$ clusters, with each cluster corresponding to one type pair for the relation.
The cluster labels are regarded as type pairs themselves.
Following this, the similarity scores of the type pairs are calculated, and the ones with the highest similarity are considered as candidate pairs to be merged together and placed in a single cluster.
To generate the `ground truth', the data points (\mbox{\boldmath$\Delta_{r}$} vectors) corresponding to both type pairs are assigned the same distinct label.
The number of clusters is given as $\mathcal{L}$ - 1 during the next iteration of clustering, and the cluster labels consist of  $\mathcal{L}$ - 2 original type pairs and one merged type pair.
If two combinations of type pairs have the same similarity score in any iteration, ties are broken arbitrarily.
This process of selecting the most similar pair of class combinations for lowering the number of clusters and obtaining cluster labels as ground truth is repeated until all type pairs have been gradually merged back together in a single cluster.
For each iteration, the quality of the clusters is evaluated and compared to the engineered ground truth, and the best performing cluster configuration is chosen as the optimal set of clusters $\mathcal{C}_{opt}$ that best represent the sub-relations with fine-grained semantics\footnote{It is to be noted that the labeling of newly proposed sub-relations is a separate task on its own. In this work, we merely utilize names of entity types to derive basic labels, however a proper naming scheme for these relations is a complex task in the context of ontology design and out of the scope of the current work.}.

\section{Experimental Analysis}
\label{sec:evaluation}
\input{tables/cluster_scores}

To evaluate the performance of the \textit{FineGReS}, we performed intrinsic empirical analysis in terms of the quality of the optimal clusters as obtained from the method. Further, to show the performance gains from \textit{FineGReS} for a relevant use case, we performed extrinsic evaluation for the task of entity classification.

\paragraph{Datasets.}
We prepared datasets derived from Yago3 and NELL-995 knowledge graphs for relation analysis and disambiguation in this work. 
The Yago dataset consists of 1,492,078 triples, with 31 relations and 917,325 unique entities. The NELL dataset includes 154,213 triples, with 200 relations and 75,492 unique entities. For both, the entities were augmented with their types as derived from the respective ontologies of the KGs.

\paragraph{Vectors for Entity Types.} 
In order to obtain word vector representations of the class types, we use the pre-trained ConVec embeddings~\cite{NLDBSherkat2017}.
We also leveraged the pre-trained \textit{Sentence-BERT}~\cite{reimers2019sentence} models from the HuggingFace library~\cite{wolf2019huggingface}. 
Cosine similarity measure was used for calculating the vector similarities (Euclidean similarity measure provided very similar results).

\paragraph{Knowledge Graph Embeddings.}
We perform our experiments on the following widely used KG embedding models : TransE~\cite{bordes2013translating} and DistMult~\cite{yang2014embedding}. These models are chosen to serve as prominent examples of embeddings using translation distance and semantic matching techniques respectively. We use the model implementations from the LibKGE library~\cite{libkge} for the Yago3-10 dataset and from OpenKE library~\cite{han2018openke} for the NELL-995 dataset. It is important to note that in this work, we have purposefully chosen those embedding techniques that showed the promise of being able to differentiate the multiple semantics of a relation. Indeed there have been some embedding models~\cite{xiao2016transg} that instead aim to encapsulate the different semantics of the relation in a single representative vector, however such techniques are not fruitful towards our goal of fine-grained relation refinement.

\input{tables/example_results}

\paragraph{Baselines.}
We establish several baselines to evaluate and compare the performance of the optimal clusters (hence the corresponding sub-relations) derived from \mbox{\textit{FineGReS}} to naive approaches in each experimental setting. \\
\noindent\textit{\textbf{max}} - Sub-relations are obtained on the basis of every different type pair that is found associated with a relation, this naive setting corresponds to the maximum number of clusters.\\
\noindent\textit{\textbf{head}} - Sub-relations 
represent all type pairs associated with a common head entity and different tail entities. It is equivalent to grouping the type pairs in the \textit{max} setting by head entity types.\\
\noindent\textit{\textbf{tail}} - Similar to \textit{head}, just replacing head entity with tail entity instead so that sub-relations represent all type pairs associated with a common tail entity and different head entities.

\paragraph{Clustering Techniques.} 
To explore the affect of different techniques during the clustering step of \mbox{\textit{FineGReS}} method, we employed several algorithms : KMeans clustering(KMC), Spectral (SPC), Optics (OPC) and Hierarchical Agglomerative clustering(HAC). 
\subsection{Cluster Quality Evaluation}

The quality of the clusters, and thereby, the resultant sub-relations is measured in terms of homogeneity score~\cite{jain2018type}, this metric favors a clustering scheme where every cluster represents a unique dominant label that corresponds to one or more unique type pairs in our case. Thus, this metric best represents the distinct semantics of the clusters. We measure and report the weighted (as per the number of data points) homogeneity scores for the clusters of all relations in the different settings of KG embeddings and clustering techniques in Table~\ref{tab:cluster-scores}.
It can be seen that in most cases the $\mathcal{C}_{FGReS}$ clusters obtained by the proposed \mbox{\textit{FineGReS}} method obtain higher scores than all the other baselines that were considered for defining the cluster configurations for sub-relations. This indicates the efficacy of the method for finding optimal fine-grained sub-relations.%

\paragraph{Discussion.}

Table~\ref{tbl:samplessplits} shows a few representative examples of the sub-relations obtained by \mbox{\textit{FineGReS}} in different settings for Yago and NELL. It can be seen that semantically different entity type pairs have been clearly separated out as distinct sub-relations, e.g. the $\langle sovereign, building \rangle$ pair for \textit{owns} relation where \textit{sovereign} is semantically distant from other types or \textit{agentCompetesWith} where $\langle bank, bank \rangle$ is a separate sub-relation. Other sub-relations have multiple type pairs associated with them based on their semantic proximity. Note that in a few cases, the optimal configuration could indeed be the \textit{max} or \textit{head}/\textit{tail} setting for the relation depending on the associated type pairs. The \textit{FineGReS} method is able to automatically determine this optimal configurations of the sub-relations for each relation relying solely on the facts in the dataset and the associated type information.

\subsection{Entity Classification Use Case}

\input{tables/entity_typing}

In order to empirically evaluate the \textit{FineGReS} method in terms of the usefulness of the derived sub-relations, we consider the popular use case of entity classification which is an important task for KG completion~\cite{neelakantan2015inferring}. 
It is modeled as a supervised multi-label classification task, where the entities are assigned to their respective types. 
Previous works have performed type prediction for entities in KGs based on statistical features~\cite{paulheim2013type}, textual information~\cite{kliegr2016lhd} as well as embeddings~\cite{biswas2020entity}. Taking cue from the same, we built a CNN classifier~\cite{zhang2017sensitivity} for the multi-label classification task which can jointly classify both the entities in a given triple to their respective types.

The dataset for the classification task was obtained by replacing the original polysemous relations in the KG dataset with their corresponding fine-grained sub-relations in the affected triples, obtained from the best performing setting of the \mbox{\textit{FineGReS}} method as well as from the baseline techniques described in Section~\ref{sec:evaluation}. The performance of entity classification measured in terms of weighted precision, recall and F1 scores (averaged over 10 runs) is shown in Table~\ref{tab:entity-typing}.
It can be seen that with well-defined relation semantics, the performance of the entity classification task improved considerably. 
In particular, the gains seen over the \textit{max} setting are indicative of the superiority of the \textit{FineGReS} method in terms of not merely finding \textit{any} set of sub-relations but finding the \textit{optimal} configuration of the sub-relations that best represent fine-grained semantics for all relations. (Similar results were also obtained for the NELL dataset).

\section{Related Work}

There is a large body of work in linguistics that deals with multiple semantics of words~\cite{erk2008structured,reisinger2010multi, neelakantan2014efficient}. 
Early work on the semantic connections between the relations and their associated entities in texts was introduced in 1963~\cite{katz1963structure} with the concept of \textit{selectional preference} for performing predicate or verb sense disambiguation~\cite{resnik1997selectional}. The idea advocates that verbs can semantically restrict the types of the arguments that occur around them, thus having a preference for certain classes of entities. %
In contrast, our work in this paper focuses on the predicates that are generic and have 
insufficient constraints for their entity types. In such cases, the diverse entity types were, in fact, leveraged to identify and define the fine-grained semantics of these predicates by dividing them into multiple predicates.

In other work related to relation semantics, ~\cite{jiang2020learning} explores the entailment between relations, e.g. the relation \textit{creator} entails \textit{author} or \textit{developer} in the sense that \textit{creator} subsumes the other relations. Similar to our work, the authors leverage the entity type information to solve the multi-classification problem of assigning the child relations to the parent ones. Our problem statement of fine-grained relation refinement is significantly more challenging and impactful in the sense that it involves the identification of novel sub-relations in an unsupervised manner.

While the idea of learning better embeddings for \textit{words} by considering their multiple contextual semantics is not new~\cite{vu2016k}, the semantics of \textit{relations} have also been recently studied in regards to learning knowledge graph embeddings.
In ~\cite{lin2015learning} the authors 
advocated the need for learning multiple relation vectors to capture the fine-grained semantics, however this study was limited in scope and lacked any consideration for complex entity type hierarchies in KGs.
In~\cite{zhang2018knowledge}, the authors create a 3-level relation hierarchy which combines similar relations as well splits relations into sub-relations, in order to improve the embeddings for relations. The proposed approach is quite rigid and opaque in terms of the actual semantics of the relations obtained from it.
In fact, the number of clusters was predefined for all relations across a dataset, in contrast to the \textit{FineGReS} method that can determine an optimal number of clusters separately for each relation based on the associated entity types.

The diverse semantics of relations was also considered by~\cite{ji2015knowledge} where the authors proposed two different vectors for the relations as well as entities, to capture their meanings and connections with each other. Similarly, in ~\cite{xiao2016transg} the authors discussed the generation of multiple translation components of relations based on their semantics with the help of a bayesian non-parametric infinite mixture model. However, they do not perform a systematic analysis of the relations semantics and a qualitative evaluation of their approach is missing.  %

In general, the above discussed works have leveraged relation polysemy for designing better embedding models and improving their link prediction performance on knowledge graphs.
In this work, we instead approach the problem of relation polysemy and discovery of the latent relation semantics with the goal of knowledge graph refinement and improvement of the quality of the relations in underlying ontology. 
More importantly, none of the previous works have explored the challenges of deriving fine-grained relations from an existing polysemous relation in the presence of complex semantic relationships between the associated entity types, which is quite common for real-world datasets. Our proposed \textit{FineGReS} method performs this task in a systematic and data-driven fashion and shows promising benefits for downstream applications.

\section{Conclusion}
In this paper, we have studied the need for relation disambiguation for knowledge
graphs due to the inherent relation polysemy in these datasets. We have proposed
a scalable, data-driven method \textit{FineGReS} that automatically determines an opti-
mal configuration for deriving sub-relations with concrete semantics. Empirical evaluation has demonstrated the efficacy of the method for learning fine-grained relation semantics for real-world data. The performance improvement achieved for downstream application of entity classification strongly indicates the promise of this approach. 
Since the method relies on the type information of the entities, \textit{FineGReS} can currently be applied only to the KGs accompanied by their ontologies, it would be interesting to extend the proposed approach to derive entity semantics from other sources, such as text.
As future work, we also plan to perform a systematic analysis of the utility and impact of this method on further NLP tasks, such as relation extraction and question answering over KGs.

\bibliography{bibliography}
\bibliographystyle{acl_natbib}

\end{document}

%% file: figures/relations_plots.tex
\begin{figure}
\centering
\includegraphics[width=1.1\columnwidth]{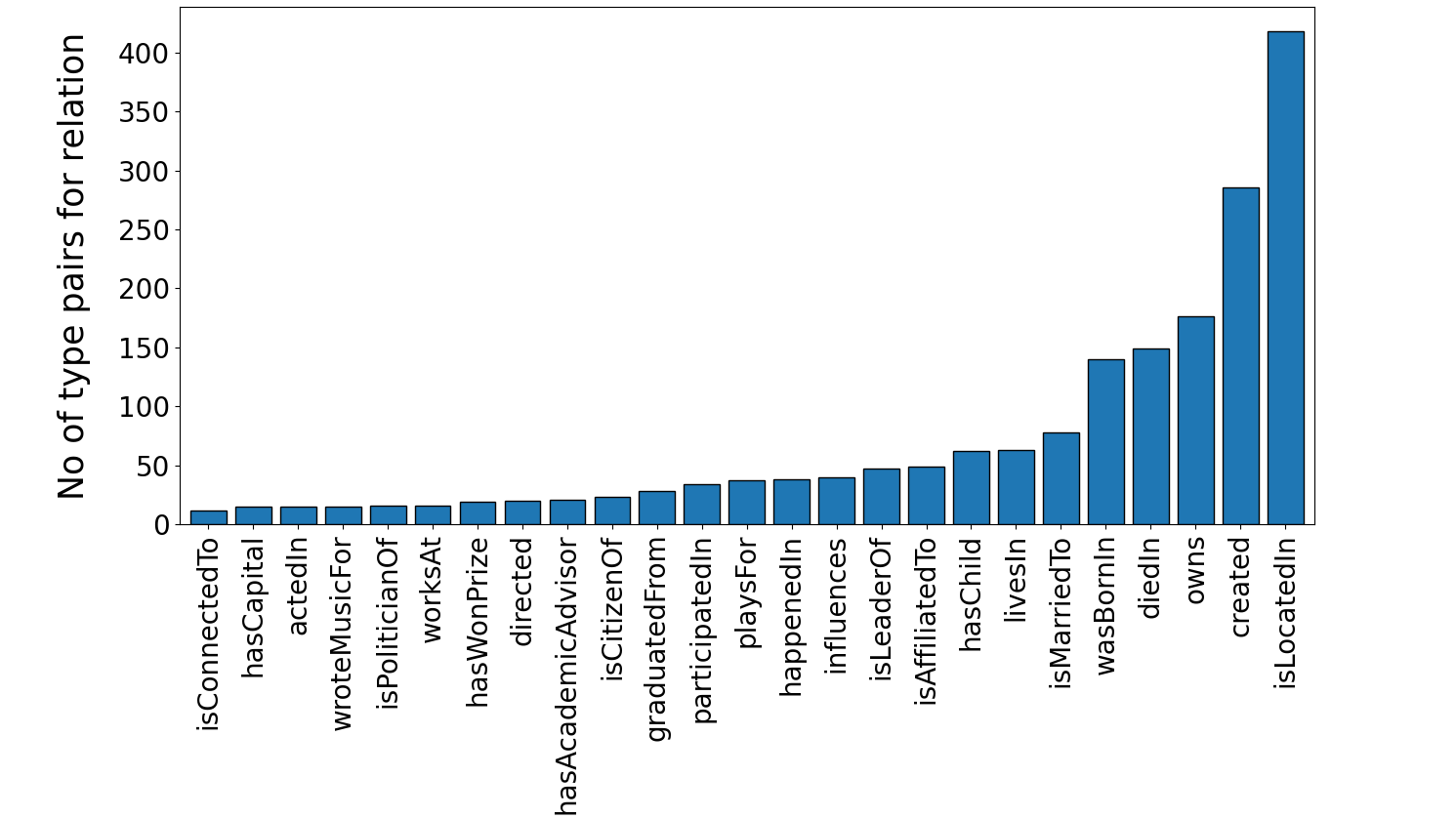}
\caption{The type pairs associated with different relations in Yago.} \label{fig:YagoRelationClassTypes}
\end{figure}

%% file: tables/examplesemantics.tex
\begin{table}[!t]
\caption{Examples of Multiple Semantics of Relations.}
\label{tbl:sampletypes}
\centering
\begin{adjustbox}{width=1\columnwidth}
\begin{tabular}{@{}c|c@{}}
\toprule
\begin{tabular}[x]{l} \textbf{Yago} \textbf{\textit{created}}\\  \end{tabular} & \begin{tabular}[x]{l} \textbf{NELL} \textit{\textbf{agentBelongsTo-}} \\ \textit{\textbf{Organization}} \end{tabular} \\
\midrule
(writer, movie) & (politician, politicalparty)  \\
(player, movie) &  (country, sportsleague) \\
(artist, movie)  & (sportsteam, sportsleague) \\
(officeholder,movie) & (coach---sportsleague) \\ 
(writer,fictional\_character) &  (person, charactertrait)\\
(artist,computer\_game) & (televisionstation, company) \\
(artist,medium)     &   \\
(writer,television)  &  \\
(company,computer\_game) &\\
\bottomrule
\end{tabular}
\end{adjustbox}
\end{table}

%% file: figures/vector_plots.tex
\begin{figure}[tb]
\centering
\includegraphics[width=0.8\columnwidth]{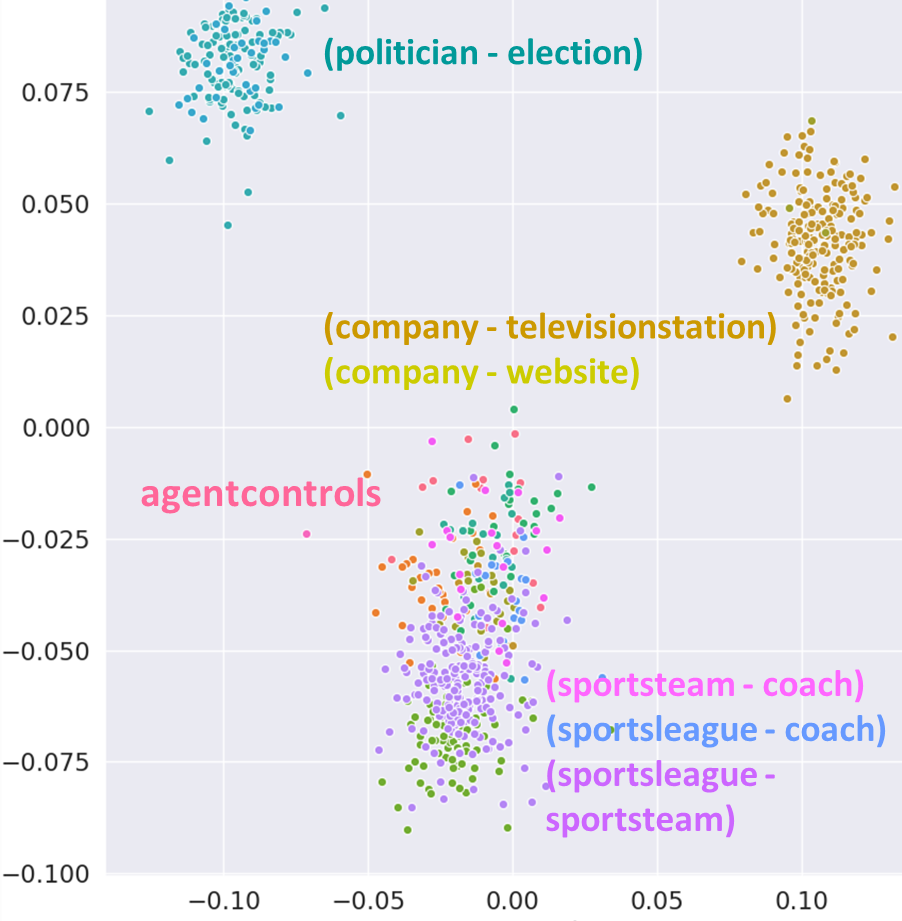}
\caption{Visualization of vectors for \textit{agentcontrols} relation in NELL with associated type pairs.} \label{fig:nell-vectors}
\end{figure}

%% file: tables/cluster_scores.tex
\begin{table*}[!htb]
\caption{Quality of \textit{FineGReS} clusters ($C_{FGReS}$) in comparison with baselines.}
\label{tab:cluster-scores}
\centering
\begin{adjustbox}{width=0.8\textwidth}
\begin{tabular}{@{}cl|r|r|r|r|r|r|r|r@{}}
\toprule
\multicolumn{2}{l}{}  & \multicolumn{4}{c}{Yago} & \multicolumn{4}{c}{NELL} \\ \midrule
 & \multicolumn{1}{c|}{\begin{tabular}[c]{@{}c@{}}Clustering \\ Technique\end{tabular}} & $C_{max}$ & $C_{head}$ & $C_{tail}$ & $C_{FGReS}$ & $C_{max}$ & $C_{head}$ & $C_{tail}$ & $C_{FGReS}$\\ \midrule
\multirow{4}{*}{TransE}
 & KMC & .245 & .199 & .190 & \textbf{.269} & .384 & .304 & .318 & \textbf{.463} \\
 & OPC & .456  & .506 & 422  & \textbf{.524}  &.192  & .197 & .203 &\textbf{.258} \\
 & SPC & \textbf{.040}  & .027 & .012 & .031 & .332  & .185 & .190 & \textbf{.337} \\
 & HAC & .217 & .195 & .182 & \textbf{.254} & .335 & .245 & .293 & \textbf{.374} \\ \cmidrule(r){1-10}
\multirow{4}{*}{DistMult}
 & KMC & \textbf{.186} & .101 & .166 & .183 & .348  & .212 & .298 & \textbf{.369}  \\
 & OPC  & .430 & .424 & .423 & \textbf{.451 }& .342 & 340 & .349  & \textbf{.370} \\
 & SPC & .316 & \textbf{.469} & .031 & .332 & .287  & .163  & .196 & \textbf{.307} \\
 & HAC & .212 & .208 & .176 & \textbf{.237}  & .283 & .173  & .227 & \textbf{.292} \\ 
\bottomrule
\end{tabular}
\end{adjustbox}
\end{table*}

%% file: tables/example_results.tex
\begin{table*}[!htb]
\caption{Examples of Fine-grained Sub-relations.}
\label{tbl:samplessplits}
\centering
\begin{adjustbox}{width=0.95\textwidth}
\begin{tabular}{@{}l|c@{}}
\toprule
\begin{tabular}[x]{l} Dataset - Relation  \\ (setting)\end{tabular} &   \textit{FineGReS} Sub-relations \\

\midrule
\begin{tabular}[x]{l} Yago - \textit{owns} \\ (TransE-HAC)\end{tabular} &
\begin{tabular}[x]{c}\{$\langle company, airport \rangle$ $\langle organization, airport \rangle$\}, \{$\langle sovereign, building \rangle$\}, \\ 
\{$\langle company, club \rangle$ $\langle company, company \rangle$, $\langle country, club \rangle$ \} \\ 
\end{tabular}\\
\midrule

\begin{tabular}[x]{l} Yago - \textit{created} \\ (TransE-OPC)\end{tabular} &
\begin{tabular}[x]{c}\{$ \langle artist, medium \rangle$ $ \langle officeholder, movie \rangle$\}, \{$ \langle writer, fictional\_character \rangle$\}, \\ \{$ \langle writer, movie \rangle$ $ \langle writer, television \rangle$ $ \langle writer, fictional\_character \rangle$ $ \langle artist, movie \rangle$\\ $ \langle artist, computer\_game \rangle$ $ \langle player, movie \rangle$\}, \{$ \langle company, computer\_game \rangle$\}
\end{tabular}\\
\midrule
\begin{tabular}[x]{l} NELL-\textit{agentCompetesWith} \\ (TransE-Kmeans)\end{tabular} &
\begin{tabular}[x]{@{}c@{}}\{$ \langle company, person \rangle$ $ \langle website, person \rangle$ $ \langle person, person \rangle$, $ \langle sportsteam, sportsteam \rangle$\}, \\
\{$ \langle person, company \rangle$, $ \langle person, website \rangle$\} \{$ \langle animal, animal \rangle$, $ \langle bird, animal \rangle$\}, \\ \{$ \langle bank, bank \rangle$\}, \{$ \langle mammal, politicsissue \rangle$\}\end{tabular}\\
\midrule
\begin{tabular}[x]{l} NELL-\textit{subpartOfOrganization} \\ (DistMult-Kmeans)\end{tabular} &
\begin{tabular}[x]{@{}c@{}}\{$\langle sportsteam, sportsteam \rangle$ $\langle stateorprovince, sportsteam \rangle$ $\langle university, sportsteam \rangle$ \\ $\langle city, sportsteam \rangle$ \}, 
  \{$\langle organization, organization \rangle$\}, \{$\langle televisionstation, city \rangle$\}, \\ \{$\langle company, company \rangle$ $\langle televisionstation, company \rangle$\}, 
\{$\langle sportsteam, sportsleague \rangle$\}, \\ \{$\langle bank, bank \rangle$\}  \{$\langle televisionstation, televisionnetwork \rangle$\}, \{$\langle televisionstation, website \rangle$\}
\end{tabular}\\
\bottomrule
\end{tabular}
\end{adjustbox}
\end{table*}

%% file: tables/entity_typing.tex
\begin{table}[]
\caption{Performance Comparison for Entity Classification Task for Yago (\textit{r} refers to original relations).}
\label{tab:entity-typing}
\begin{adjustbox}{width=0.48\textwidth}
\begin{tabular}{@{}l|c|r|r|r|r|r}
\toprule
 & \textit{r} & \textit{max} & \textit{head} & \textit{tail} & \multicolumn{2}{c}{\textit{FineGReS}}\\ 
&                    &                      &                       &                       & \multicolumn{1}{l|}{TransE} & DistMult \\ \midrule
Precision         & .893               & .916                 & .906                  & .322                  & \multicolumn{1}{r|}{\textbf{.923}}   & \textbf{.928}     \\ 
Recall            & .908               & .925                 & .921                  & .425                  & \multicolumn{1}{r|}{\textbf{.941}}   & \textbf{.942}     \\ 
F1 Score          & .894               & .914                 & .909                  & .34                   & \multicolumn{1}{r|}{\textbf{.931}}   & \textbf{.931}     \\ \bottomrule
\end{tabular}
\end{adjustbox}
\end{table}